\documentclass{article}
\usepackage{amsmath,graphicx,mlspconf}
\usepackage{amssymb}
\usepackage{mathtools}
\usepackage{subfigure}
\usepackage{algorithmic}
\usepackage{algorithm}
\usepackage{bbm}
\usepackage{cite}
\usepackage{url}

\copyrightnotice{U.S.\ Government work not protected by U.S.\ copyright}

\copyrightnotice{978-1-4673-1026-0/12/\$31.00 {\copyright}2013 Crown}

\copyrightnotice{978-1-4799-1180-6/13/\$31.00 {\copyright}2013 IEEE}

\toappear{2013 IEEE INTERNATIONAL WORKSHOP ON MACHINE LEARNING FOR SIGNAL PROCESSING, SEPT.\ 22--25, 2013, SOUTHAMPTON, UK}

\def\argmax{\arg\!\max}

\title{NOVELTY DETECTION UNDER MULTI-INSTANCE MULTI-LABEL FRAMEWORK}
\name{Qi Lou, Raviv Raich, Forrest Briggs and Xiaoli Z. Fern\thanks{This work was partially supported by the
National Science Foundation grant CCF-1254218.
}}
\address{School of EECS, Oregon State University, Corvallis, OR 97331-5501, USA\\
         \{lou, raich, briggsf, xfern\}@eecs.oregonstate.edu}
\begin{document}
%

\maketitle

\begin{abstract}
Novelty detection plays an important role in machine learning and signal processing.
This paper studies novelty detection in a new setting
where the data object is represented as a bag of instances and associated with multiple class labels,
referred to as multi-instance multi-label (MIML) learning.
Contrary to the common assumption in MIML that each instance in a bag belongs to one of the known classes,
in novelty detection, we focus on the scenario where bags may contain novel-class instances.
The goal is to determine, for any given instance in a new bag, whether it belongs to a known class or a novel class.
Detecting novelty in the MIML setting captures many
real-world phenomena and has many potential applications. For example, in a collection of tagged images,
the tag may only cover a subset of objects existing in the images. Discovering an object whose class has not been previously tagged
can be useful for the purpose of soliciting a label for the new object class.
To address this novel problem, we present a discriminative framework for detecting new class instances.
Experiments demonstrate the effectiveness of our proposed method,
and reveal that the presence of unlabeled novel instances in
training bags is helpful to the detection of such instances in testing stage.
\end{abstract}
\section{INTRODUCTION}
\label{sec:intro}
Novelty detection is the identification of new or unknown data
that is not labeled
during training \cite{Markou03_1}.  In the traditional setting,
only training examples from a nominal distribution are provided and the
goal is to determine for a new example whether it comes from the nominal
distribution or not. Much work has been
done in this field. Early work is generally divided into
two categories \cite{Markou03_1, Markou03_2}.
One category includes statistical approaches such as some density estimation methods.
The other category consists of neural network based approaches, e.g., multi-layer perceptrons.
Several new approaches have been introduced in recent years.
In \cite{Hero06}, geometric entropy minimization is introduced for anomaly detection.
An efficient anomaly detection method using bipartite $k$-NN graphs is presented in \cite{Sricharan11}.
In \cite{Zhao09}, an anomaly detection algorithm is proposed
based on score functions. Each point gets scores
from its nearest neighbors. This algorithm
can be directly applied to novelty detection.
In \cite{Scholkopf99},
SVMs are applied to novelty detection to learn a
function $f$ that is positive on
a subset $S$ of the input space and negative outside $S$.

In this paper, we consider novelty detection in a new setting where the data follows a multi-instance multi-label (MIML) format.
The MIML framework has been primarily studied for supervised learning \cite{Zhou12} and widely used in applications where data is associated with multiple classes and can be naturally represented as bags of instances (i.e., collections of parts).
For example, a document can be viewed as a bag of words and associated with multiple tags.
Similarly, an image can be represented as a bag of pixels or patches,
and associated with multiple classes corresponding to the objects that it contains.
Formally speaking, the training data in MIML consists of a collection of labeled bags
$\{(X_{1}, Y_{1}),(X_{2}, Y_{2}),\ldots, (X_{N}, Y_{N})\}$,
where $X_i\subset X$ is a set of instances and $Y_i\subset Y$ is a set of labels.
In the traditional MIML applications the goal is to learn
a bag-level classifier $f: 2^X\rightarrow 2^Y$ that can reliably predict the label set of a previously unseen bag.

It is commonly assumed in MIML that every instance we observe in the training set belongs to one of the known classes. However, in many applications, this assumption is violated. For example, in a collection of tagged images, the tag may only cover a subset of objects present in the images.
The goal of novelty detection in the MIML setting is to determine whether a given instance comes from an unknown class given only a set of bags labeled with the known classes.
This setup has several advantages compared to the more well-known setup in
novelty detection: First, the labeled bags allow us to apply an approach that takes into account the presence of multiple known classes.
Second, frequently the training set would contain some novel class instances. The presence of such instances, although never explicitly labeled as novel instances, can in a way serve as ``implicit'' negative examples for the known classes, which can be helpful for identifying novel instances in new bags.

The work presented in this paper is inspired by a real world bioacoustics application.
In this application, the annotation of individual bird vocalization is often a time consuming task.
As an alternative, experts identify from a list of focal bird species the ones that they recognize in a given recording.
Such labels are associated with the entire recording and not with a specific vocalization in the recording.
Based on a collection of such labeled recordings,
the goal is to annotate each vocalization in a new recording \cite{Briggs12}.
An implicit assumption here is that each vocalization in the recording must come from one of the focal species, which can be incomplete.
Under this assumption, vocalizations of new species outside of the focal list will not be discovered.
Instead, such vocalizations will be annotated with a label from the existing species list.
The setup proposed in this paper allows for novel instances to be observed in the training data without being explicitly labeled,
and hence should enable the annotation of vocalizations from novel species.
In turn, such novel instances can be presented back to the experts for further inspection.

To the best of our knowledge, novelty detection in the MIML setting has not
been investigated. Our main contributions are: (i) We propose
a new problem -- novelty detection in the MIML setting.
(ii) We offer a framework based on score functions to solve the problem.
(iii) We illustrate the efficacy of our method on a real-world MIML bioacoustics data.

\section{PROPOSED METHODS}
\label{sec:methods}

Suppose we are given a collection of labeled bags
$\{(X_{1}, Y_{1}),$ $(X_{2}, Y_{2}),$  $\ldots,$
$(X_{N}, Y_{N})\}$, where the $i$th bag $X_i\subset X$ is
a set of instances from the feature space
$X\subset \mathbb{R}^d$, and $Y_{i}$ is a subset of the
know label set $Y=\bigcup^{N}_{i=1}Y_{i}$.
For any label $y_{im}\in Y_{i}$, there is at least
one instance $x_{in}\in X_{i}$ belonging
to this class. We consider the scenario where an instance
in $X_{i}$ has no label in $Y_{i}$ related to it,
which extends the traditional MIML learning framework.
Our goal is to determine for a given instance $x\in X$
whether it belongs to a known class in $Y$ or not.

To illustrate the intuition behind our general strategy,
consider the toy problem shown in Table~\ref{tb:toy_problem}.
The known label set is \{\uppercase\expandafter{\romannumeral1},\uppercase\expandafter{\romannumeral2}\}.
We have four labeled bags available.
According to the principle that one instance must belong to one class and one bag-level label must
have at least one corresponding instance, we conclude that
$\vartriangle$ is drawn from class \uppercase\expandafter{\romannumeral1}, $\square$ belongs
to class \uppercase\expandafter{\romannumeral2}, and $\lozenge$ doesn't come from the existing
classes. $\triangledown$ cannot be fully determined based on
current data.

\begin{table}[htb]
\caption{Toy problem with two known classes}
\label{tb:toy_problem}
\begin{center}
\begin{tabular}{| c | c | c |}
\hline
$\#$ & Bags ($X_i$) & Labels ($Y_i$) \\ \hline
$1$ & $\{\vartriangle\vartriangle\square\triangledown\}$ & \{\uppercase\expandafter{\romannumeral1}, \uppercase\expandafter{\romannumeral2}\} \\\hline
$2$ & $\{\vartriangle\vartriangle\lozenge\triangledown \}$ & \{\uppercase\expandafter{\romannumeral1}\}  \\\hline
$3$ & $\{\square\square\square\lozenge\lozenge\}$ &  \{\uppercase\expandafter{\romannumeral2}\} \\\hline
$4$ & $\{\vartriangle\vartriangle\square\} $ & \{\uppercase\expandafter{\romannumeral1}, \uppercase\expandafter{\romannumeral2}\} \\\hline
\end{tabular}
\end{center}
\end{table}

\begin{table}[htb]
\caption{Concurrence rates for the toy problem}
\label{tb:co-occurrence}
\begin{center}
\begin{tabular}{|l|*{5}{c|}}
\hline
  & $\vartriangle$ & $\triangledown$ & $\square$ & $\lozenge$ \\ \hline
\uppercase\expandafter{\romannumeral1} &  $1$  & $3/4$  & $1/2$  & $1/4$    \\\hline
\uppercase\expandafter{\romannumeral2} & $1/2$ & $1/4$  &  $1$   & $1/4$    \\\hline
\end{tabular}
\end{center}
\end{table}

To express this observation mathematically, we calculate
the rate of co-occurrence of an instance
and a label. For example, $\vartriangle$ appears with
label \uppercase\expandafter{\romannumeral1} together in bag $1$, $2$, $4$ and
they are both missing in bag $3$. So, the co-occurrence
rate $p(\vartriangle,$ \uppercase\expandafter{\romannumeral1}$) = 1$.
All the other rates are listed in Table~\ref{tb:co-occurrence}.
If we detect an instance based on the maximal co-occurrence rate with
respect to all classes and set a threshold to be $3/4$,
we will reach a result that can generally reflect our previous observation.

This example inspires us to devise a
general strategy for detection. We introduce a set of score functions,
each of which corresponds to one class, i.e., for each
label $c\in Y$, we assign a function $f_c$ to class $c$.
Generally, for an instance from a specific known class,
the value of the score function corresponding to this class
should be large. If all scores of an instance
are below a prescribed threshold, it would not be considered
to belong to any known class. The decision principle is:
If $\max_{c \in \{1, \ldots,|Y|\}} f_c(x) < \varepsilon$ then return `unknown', otherwise return `known'.

There are many possible choices for the set of score functions.
Generally, the score functions are expected to enable us to
achieve a high true positive rate  with a given
false positive (Type \uppercase\expandafter{\romannumeral1} error) rate, which
can be measured by the area under the curve (AUC) of ROC.

\subsection{Kernel Based Scoring Functions}
\label{sec:algorithm}
We define the score function for class $c$ as follows:
\begin{equation}\label{eq:kernel}
\begin{array}{rcl}
f_c(x)&=&\sum\limits_{x_l\in\underset{i}{\bigcup}X_i}\alpha_{cl}k(x,x_l) \\
&=& \alpha_c^\mathrm{T}k(x)
\end{array}
\end{equation}
where $X_i$'s are training bags, $x_l$'s are training instances from training bags,
$k(\cdot,\cdot)$ is the kernel function such that $k(x)=(k(x,x_1),\ldots,k(x,x_L))^\mathrm{T}$,
and $\alpha_{cl}$'s are the components of the weight vector $\alpha_c=(\alpha_{c1},\ldots,\alpha_{cL})^\mathrm{T}$.


We encourage $f_c$ to take positive values on instances
in class $c$ and negative values on instances from other classes.
Hence, we define the objective function $\mathcal{OBJ}$ as
\begin{equation}
\frac{\lambda}{2}\sum\limits_{c=1}^{|Y|}\alpha_c^\mathrm{T}\mathbf{K}\alpha_c +
\frac{1}{N|Y|}\sum\limits_{i=1}^{N}\sum\limits_{c=1}^{|Y|}F_{c}(X_{i})
\end{equation}
where
\[F_{c}(X_{i})=\max\{0, ~1-y_{ic}\max\limits_{x_{ij}\in X_i}f_c(x_{ij})\}, ~y_{ic}\in\{-1,+1\}\]
$\lambda$ is a regularization parameter,
$\mathbf{K}$ is the kernel matrix with $(i,j)$-th entry $k(x_i,x_j), x_i, x_j\in\underset{k}{\bigcup}X_k$, and
$y_{ic}=+1$ if and only if $Y_i$
contains the label for class $c$.

In fact, we define an objective function for each class separately
and sum over all these objective functions to construct $\mathcal{OBJ}$.
The first term of $\mathcal{OBJ}$ controls model complexity.
$F_c(\cdot)$ in the second term of  $\mathcal{OBJ}$
can be viewed as a bag-level hinge loss for class $c$,
which is a generalization of the single-instance case.
If $c$ is a bag-level label of bag $X_{i}$,
we expect $\max\limits_{x_{ij}\in X_i}f_c(x_{ij})$
to give a high score because there is at least
one instance in $X_{i}$ is from class $c$.
Other loss functions such as rank loss \cite{Briggs12}
have already been introduced for MIML learning.

Our goal is to minimize the objective function which is unfortunately non-convex.
However, if we fix the term $\max\limits_{x_{ij}\in X_i}f_c(x_{ij})$, i.e.,
find the support instance $x_{ic}$ such that $x_{ic}=\argmax_{x_{ij}\in X_i}{\alpha_c}^\mathrm{T}k(x_{ij})$
and substitute back to the objective function,
the resulted objective function $\mathcal{OBJ}_{\ast}$ will be convex with respect to $\alpha_c$'s.
To solve this convex problem, we deploy the L-BFGS \cite{Byrd94}
algorithm. The subgradient along $\alpha_c$ used in L-BFGS is computed as follows:
\begin{equation}
\label{eq:subgradient}
\nabla_c = \lambda\mathbf{K}\alpha_c-\frac{1}{N|Y|}\sum\limits_{i=1}^{N}y_{ic}k(x_{ic})\mathbbm{1}_{\{1-y_{ic}f_c(x_{ic})>0\}}
\end{equation}
Details can be found in Algorithm~\ref{al:descent}.
This descent method can be applied to any choice of kernel function
and according to our experience it works very well (usually converges within $30$ steps).
Note that many algorithms \cite{Briggs12,Yakhnenko11Multi-Instance} for MIML learning that attempt to
learn an instance-level score functions
including the proposed approach are based on a non-convex objective.
Consequently, no global optimum is guaranteed.
To reduce the effect induced by randomness,
we usually rerun the algorithm multiple times with independent random initializations
and adopt the result with the smallest value of the objective function.

\begin{algorithm}[htb]
\caption{Descent Method}
\label{al:descent}
\begin{algorithmic}
\REQUIRE  $\{(X_{1}, Y_{1}), (X_{2}, Y_{2}), \ldots, (X_{N}, Y_{N})\}$, $\lambda$, $T$.
\STATE Randomly initialize all $\alpha_c$'s s.t. $\|\alpha^1_c\|=1$
\FOR{$t = 1$ to $T$}
\STATE Set $x^t_{ic}=\argmax_{x_{ij}\in X_i}{(\alpha^t_c)}^\mathrm{T}k(x_{ij})$,  \\
           \hspace{6mm} $\mathbbm{1}^t_{ic}=\mathbbm{1}_{\{1-y_{ic}f_c(x^t_{ic})>0\}}$,  \\
           \hspace{6.3mm} $\nabla^t_c=\lambda\mathbf{K}\alpha_c-\frac{1}{N|Y|}\sum\limits_{i=1}^{N}y_{ic}k(x^t_{ic})\mathbbm{1}^t_{ic}$. \\
\STATE Plug $\{x^t_{ic}\}$ into $\mathcal{OBJ}$ to get a convex surrogate $\mathcal{OBJ}^t_{\ast}$.
\STATE Run L-BFGS with inputs $\mathcal{OBJ}^t_{\ast}$, $\nabla^t_c$ to return $\{\alpha^{t+1}_c\}$ and $\mathcal{OBJ}^{t+1}$
\ENDFOR
\RETURN $\{\alpha^{T+1}_c\}$ and $\mathcal{OBJ}^{T+1}$.
\end{algorithmic}
\end{algorithm}

\subsection{Parameter Tuning}
In our experiment, we use Gaussian kernel,
i.e., $k(x_i,x_j)=e^{-\gamma\|x_i-x_j\|^2}$, where $\|\cdot\|$ is the Euclidean norm.
The parameter $\gamma$ controls the bandwidth of the kernel.
Hence, there are a pair of parameters $\lambda$ and $\gamma$ in the objective function required to be determined.

While training, we search in a wide range of values for the parameter pair,
and select the pair with corresponding $\alpha_c$'s that minimizes
\[\sum\limits_{i=1}^{N}\sum\limits_{c=1}^{|Y|}g(y_{ic}\max\limits_{x_{ij}\in X_i}f_c(x_{ij}))\]
where $g(x) = \mathbbm{1}_{x<0}$ is the zero-one loss function.
Note that $\mathbbm{1}_{x<0}$ is a lower bound of the hinge loss $\max\{0,1-x\}$.

We vary the value of threshold to generate ROCs while testing.
The values of threshold are derived from training examples.

\section{EXPERIMENTAL RESULTS}
\label{sec:experiment}
In this section, we provide a number of experimental results
based on both synthetic data and real-world data to show the effectiveness of our algorithm.
Additionally, we present a comparison to one-class SVM, a notable anomaly detection algorithm.

\subsection{MNIST Handwritten Digits Dataset}
We generated the synthetic data based on the MNIST handwritten digits data
set\footnote{Available on-line \url{http://www.cs.nyu.edu/~roweis/data.html}}.
Each image in the data set is a $28$ by $28$ bitmap, i.e.,
a vector of $784$ dimensions. By using PCA, we reduced the dimension of instances to $20$.

\begin{table}[htb]
\caption{Bag examples for the handwritten digits data. We take the first four digits
`0', `1', `2', `3' as known classes, i.e., $Y=$\{`0', `1', `2', `3' \}.
In each bag, some instances are without associated labels. For example,
in bag 1 examples for `5' and `9' are considered from unknown classes.}
\label{tb:digits_bag}
\begin{center}
\begin{tabular}{|l|*{3}{c|}}
\hline
number &  bags & labels  \\ \hline
1& \includegraphics[width=5.0cm]{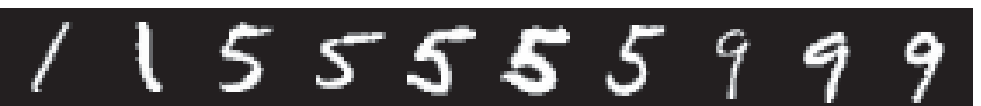}  &  {`1'}\\ \hline
2& \includegraphics[width=5.0cm]{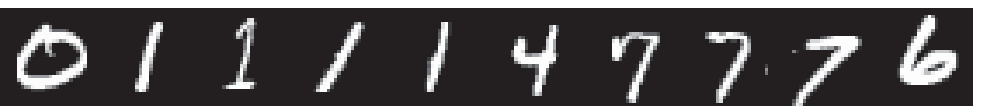}  &  {`0', `1'}\\ \hline
3& \includegraphics[width=5.0cm]{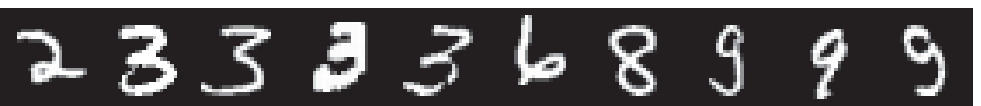}  &  {`2', `3'}\\ \hline
4& \includegraphics[width=5.0cm]{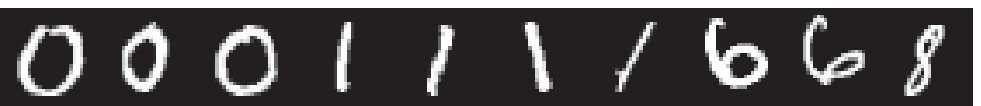}  &  {`0', `1'}\\ \hline
5& \includegraphics[width=5.0cm]{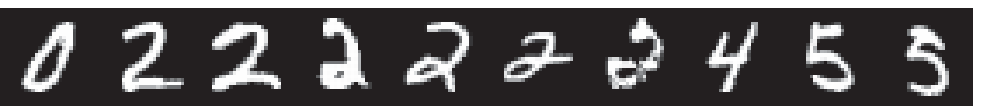}  &  {`0', `2'}\\ \hline
\end{tabular}
\end{center}
\end{table}

We created training and testing bags from the MNIST instances.
Some examples for handwritten digits bags are shown in Table~\ref{tb:digits_bag}.
Two processes for generating bags are listed in Algorithm~\ref{al:bag_generation}
and Algorithm~\ref{al:bag_generation_filter}. The only difference between these
two procedures is that Algorithm~\ref{al:bag_generation_filter} rules out the
possibility of a label set for a bag being empty, i.e., a bag including purely novel examples.
For Dirichlet process used in our simulation, we assigned relatively small concentration
parameters $\beta = (\beta_1, \beta_2, \ldots, \beta_{10})$
to the Dirichlet distribution in order to encourage a sparse label set for a bag, which is common in real-world scenarios.
We set all $\beta_i = 0.1$ and the bag size $M=20$. Typical examples of bags generated from
Dirichlet distribution are shown in Table~\ref{tb:bags}.

\begin{table}[htb]
\caption{Examples for numbers of each digit in 5 bags when each component of $\beta$ is $0.1$.
The bag size is set to be $20$.}
\label{tb:bags}
\begin{center}
\begin{tabular}{|l|*{10}{c|}} \hline
`0'  &   `1'  &   `2'  &   `3'  &   `4'  &   `5'  &   `6'  &   `7'  &   `8'  &   `9' \\ \hline
0  &  1  &  0  &  0  &  5  &  5  &  2  &  3  &  4  &  0  \\ \hline
6  &  0  &  4  &  8  &  0  &  0  &  1  &  0  &  0  &  1  \\ \hline
0  &  17  &  0  &  2  &  0  &  0  &  1  &  0  &  0  &  0  \\ \hline
0  &  0  &  1  &  0  &  2  &  16  &  1  &  0  &  0  &  0  \\ \hline
0  &  0  &  15  &  0  &  0  &  0  &  0  &  5  &  0  &  0  \\ \hline
\end{tabular}
\end{center}
\end{table}

\begin{algorithm}[htb]
\caption{Bag generation procedure for handwritten digits data}
\label{al:bag_generation}
\begin{algorithmic}
\REQUIRE  $N$, $M$, $Y$, $\beta$.
\FOR{$i = 1$ to $N$}
\STATE  Draw $M$ instances $\{x_{ij}\}$ according to the proportion given by Dirichlet ($\beta$) distribution.
\STATE  Extract labels from $x_{ij}$'s to form $Y_{i}^{'}$ and set $Y_i=Y\cap Y_{i}^{'}$.
\ENDFOR
\end{algorithmic}
\end{algorithm}

\begin{algorithm}[htb]
\caption{Bag generation procedure with filtration for handwritten digits data.}
\label{al:bag_generation_filter}
\begin{algorithmic}
\REQUIRE  $N$, $M$, $Y$, $\beta$.
\FOR{$i = 1$ to $N$}
\STATE  Set $Y_i=\emptyset$.
\WHILE{$Y_i==\emptyset$}
\STATE  Draw $M$ instances $\{x_{ij}\}$ according to the proportion given by Dirichlet ($\beta$) distribution.
\STATE  Extract labels from $x_{ij}$'s to form $Y_{i}^{'}$ and set $Y_i=Y\cap Y_{i}^{'}$.
\ENDWHILE
\ENDFOR
\end{algorithmic}
\end{algorithm}

\begin{figure}[htb]
\begin{minipage}[b]{\linewidth}
  \centering
  \centerline{\includegraphics[width=7.0cm]{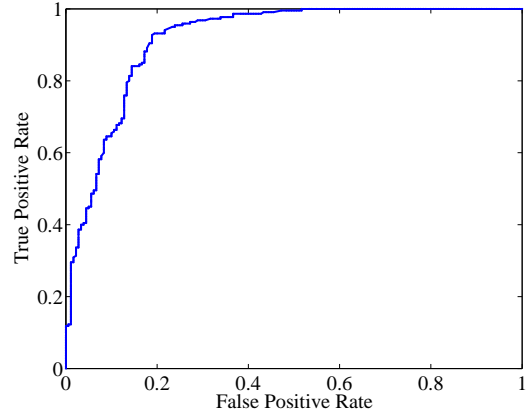}}
  \centerline{(a)}\medskip
\end{minipage}
\begin{minipage}[b]{\linewidth}
  \centering
  \centerline{\includegraphics[width=7.0cm]{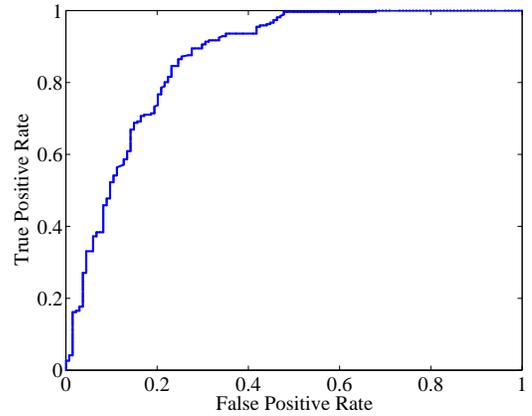}}
  \centerline{(b)}\medskip
\end{minipage}
 \vspace{-0.8cm}
\caption{Typical examples of ROCs from the handwritten digit data.
The subfigure (a) shows  a ROC example from the first setting
and the subfigure (b) gives  an example from the second setting.}
\label{fig:roc_digit}
\end{figure}

We provided our method with bags generated in two different ways:
\begin{enumerate}
\item Generate both training and testing bags according to Algorithm~\ref{al:bag_generation}.
\item Generate training bags according to Algorithm~\ref{al:bag_generation_filter} while
generate testing bags by applying Algorithm~\ref{al:bag_generation}.
\end{enumerate}
In our experiments, we consider various sizes of known label sets and different combinations of labels
in these two settings. Two typical examples of ROCs from the two setting are shown in Figure~\ref{fig:roc_digit}.

Table~\ref{tb:auc_digit} shows the average AUCs of ROCs over multiple runs from the first setting.
We observe that average AUCs are all above $0.85$ for the known label sets of size $4$.
For the known label sets of size $8$, the average AUCs are all larger than $0.8$.
The results are fairly stable with different combinations of labels. This demonstrates
the effectiveness of our algorithm. 


Table~\ref{tb:auc_digit_filter} shows the average AUCs of ROCs for the setting
which does not contain bags with an empty label set.
The label sets in these two tables are the same.
The results in the two tables are comparable but those in Table~\ref{tb:auc_digit}
are always better. This demonstrates that it is beneficial to include
bags with an empty label set. The reason could be that those bags contain
purely novel examples and hence training on those bags is very reliable.

\begin{table}[htb]
\caption{Average AUCs for handwritten digits data. Y is the known label set.
Training bags and testing bags are both generated according to Algorithm~\ref{al:bag_generation},
i.e., without bag filtration.}
\vspace{-0.4cm}
\label{tb:auc_digit}
\begin{center}
\tabcolsep=0.09cm
\begin{tabular}{|l|*{4}{c|}} \hline
$Y$  &  AUC & $Y$ & AUC \\ \hline
\{`0',`1',`3',`7'\}  & 0.89  & \{`0',`1',`2',`3',`4',`5',`6',`7'\} & 0.85  \\ \hline
\{`2',`4',`7',`8'\}  & 0.87  & \{`2',`3',`4',`5',`6',`7',`8',`9'\} & 0.88  \\ \hline
\{`2',`5',`6',`7'\}  & 0.91  & \{`0',`1',`4',`5',`6',`7',`8',`9'\} & 0.84  \\ \hline
\{`3',`5',`7',`9'\}  & 0.85  & \{`0',`1',`2',`3',`6',`7',`8',`9'\} & 0.85  \\ \hline
\{`3',`6',`8',`9'\}  & 0.89  & \{`0',`1',`2',`3',`4',`5',`8',`9'\} & 0.83  \\ \hline
\end{tabular}
\end{center}
\end{table}

 \vspace{-0.5cm}
\begin{table}[htb]
\caption{Average AUCs for handwritten digits data. Y is the known label set.
Training bags are generated according to Algorithm~\ref{al:bag_generation_filter},
i.e., with bag filtration,
while testing bags are generated by Algorithm~\ref{al:bag_generation},
i.e., without bag filtration.}
\vspace{-0.4cm}
\label{tb:auc_digit_filter}
\begin{center}
\tabcolsep=0.09cm
\begin{tabular}{|l|*{4}{c|}} \hline
$Y$  &  AUC & $Y$ & AUC \\ \hline
\{`0',`1',`3',`7'\}  & 0.86  & \{`0',`1',`2',`3',`4',`5',`6',`7'\} &  0.85  \\ \hline
\{`2',`4',`7',`8'\}  & 0.86  & \{`2',`3',`4',`5',`6',`7',`8',`9'\} &  0.84  \\ \hline
\{`2',`5',`6',`7'\}  & 0.88  & \{`0',`1',`4',`5',`6',`7',`8',`9'\} &  0.82  \\ \hline
\{`3',`5',`7',`9'\}  & 0.83  & \{`0',`1',`2',`3',`6',`7',`8',`9'\} &  0.84  \\ \hline
\{`3',`6',`8',`9'\}  & 0.86  & \{`0',`1',`2',`3',`4',`5',`8',`9'\} &  0.80  \\ \hline
\end{tabular}
\end{center}
\end{table}

\subsection{HJA Birdsong Dataset}
We tested our algorithm on the real-world
dataset - HJA birdsong dataset\footnote{Available on-line \url{http://web.engr.oregonstate.edu/~briggsf/kdd2012datasets/hja_birdsong/}},
which has been used in \cite{Briggs12Instance,Liu12}.
This dataset consists of $548$ bags, each of which contains several $38$-dimensional instances.
The bag size, i.e., the number of instances in a bag, varies vary from $1$ to $26$, the average of which is approximately $9$.
The dataset includes $4998$ instances from $13$ species.
Species names and the numbers of instances
for those species are listed in Table~\ref{tb:birdname}.
Each species corresponds to a class in the complete label set $\{1,2,\ldots,13\}$.
We took a subset of the complete label set as the
known label set and conducted experiment with various choices of the known label set.
Table~\ref{tb:auc_bird} shows the average AUCs of different known label sets.
Specifically, we intentionally made each species appear at least once in those known sets.
From Table~\ref{tb:auc_bird}, we observe that most all of the values of AUCs
are above $0.85$ and some even reach $0.9$. The results are quite stable with different
label settings despite the imbalance in the instance population of the species.
These results illustrate the potential of the approach as a utility for novel species discovery.

\begin{table}[htb]
\caption{Names of bird species and the number of total instances for each species. Each species corresponds to one class.}
\label{tb:birdname}
\begin{center}
\begin{tabular}{|l|*{3}{c|}} \hline
Class  &  Species & No. of Instances \\ \hline
1  & Brown Creeper  &  602 \\ \hline
2  & Winter Wren  &  810 \\ \hline
3  & Pacific-slope Flycatcher  & 501  \\ \hline
4  & Red-breasted Nuthatch  &  494 \\ \hline
5  & Dark-eyed Junco  & 82 \\ \hline
6  & Olive-sided Flycatcher  &  277 \\ \hline
7  & Hermit Thrush  &  32  \\ \hline
8  & Chestnut-backed Chickadee &  345 \\ \hline
9  & Varied Thrush &  139 \\ \hline
10 & Hermit Warbler & 120  \\ \hline
11 &  Swainson's Thrush  &  190 \\ \hline
12 &  Hammond's Flycatcher  &  1280 \\ \hline
13 & Western Tanager &  126 \\ \hline
\end{tabular}
\end{center}
\end{table}

\begin{table}[htb]
\caption{Average AUCs for birdsong data. Y is the known label set.}
\label{tb:auc_bird}
\begin{center}
\begin{tabular}{|l|*{4}{c|}} \hline
$Y$  &  AUC  &  $Y$  &  AUC \\ \hline
\{1,2,4,8\}   & 0.90  & \{1,2,3,4,5,6,7,8\}  & 0.89 \\ \hline
\{3,5,7,9\}   & 0.85  & \{3,4,5,6,7,8,9,10\} &  0.85 \\ \hline
\{4,6,8,10\}  & 0.88  & \{5,6,7,8,9,10,11,12\} & 0.89 \\ \hline
\{5,7,9,11\}  & 0.90  & \{1,7,8,9,10,11,12,13\} & 0.84 \\ \hline
\{6,10,12,13\} & 0.89 & \{1,2,3,9,10,11,12,13\} & 0.85 \\ \hline
\end{tabular}
\end{center}
\end{table}

\subsection{Comparison with One-Class SVM}
Our algorithm deals with detection problem with MIML setting, which is
different from the traditional setting for anomaly detection.
We argue that traditional anomaly detection algorithms cannot be directly
applied to our problem.
To make comparison, we adopt one-class SVM\cite{Scholkopf99estimating,CC01a,Manevitz01one-classsvms},
a well known algorithm for anomaly detection.
To apply one-class SVM,
we construct a normal class training data consisting of examples from the known label set.
The parameter $\nu$ vary from $0$ to $1$ with step size $0.02$ to generate ROCs.
The Gaussian kernel is used for one-class SVM.
We search the parameter $\gamma$ for the kernel in a wide range
and select the best one for on-class SMV post-hoc.
We present this unfair advantage to one-class SVM for two reasons:
(i) It is unclear how to optimize the parameter in the absence
of novel instances. (ii) We would like to illustrate the point
that even given such unfair advantage,
one-class SVM cannot outperform our algorithm.

Table~\ref{tb:auc_svm_digit} and \ref{tb:auc_svm_bird} show the average AUCs for
handwritten digits data and birdsong data respectively.
Compared to Table~\ref{tb:auc_digit} and \ref{tb:auc_bird},
the proposed algorithm outperforms 1-class SVM in terms of AUC not only in absolute value but also in stability.
This also demonstrates that training with unlabeled instances are beneficial
to the detection.


\begin{table}[htb]
\caption{Average AUCs for the handwritten digits data by applying one-class SVM with Gaussian kernel. Y is the known label set.}
\vspace{-0.4cm}
\label{tb:auc_svm_digit}
\begin{center}
\tabcolsep=0.09cm
\begin{tabular}{|l|*{4}{c|}} \hline
$Y$  &  AUC  &  $Y$  &  AUC \\ \hline
\{`0',`1',`3',`7'\}  & 0.66  & \{`0',`1',`2',`3',`4',`5',`6',`7'\} & 0.59 \\ \hline
\{`2',`4',`7',`8'\}  & 0.66  & \{`2',`3',`4',`5',`6',`7',`8',`9'\} & 0.63 \\ \hline
\{`2',`5',`6',`7'\}  & 0.57  & \{`0',`1',`4',`5',`6',`7',`8',`9'\} & 0.68 \\ \hline
\{`3',`5',`7',`9'\}  & 0.65  & \{`0',`1',`2',`3',`6',`7',`8',`9'\} & 0.62 \\ \hline
\{`3',`6',`8',`9'\}  & 0.63  & \{`0',`1',`2',`3',`4',`5',`8',`9'\} & 0.65  \\ \hline
\end{tabular}
\end{center}
\end{table}

\vspace{-0.5cm}
\begin{table}[htb]
\caption{Average AUCs for the birdsong data by applying one-class SVM. Y is the known label set.}
\label{tb:auc_svm_bird}
\begin{center}
\begin{tabular}{|l|*{4}{c|}} \hline
$Y$  &  AUC  &  $Y$  &  AUC \\ \hline
\{1,2,4,8\}   & 0.78  & \{1,2,3,4,5,6,7,8\}  & 0.85 \\ \hline
\{3,5,7,9\}   & 0.79  & \{3,4,5,6,7,8,9,10\} & 0.82 \\ \hline
\{4,6,8,10\}  & 0.82  & \{5,6,7,8,9,10,11,12\} & 0.75 \\ \hline
\{5,7,9,11\}  & 0.73  & \{1,7,8,9,10,11,12,13\} & 0.70 \\ \hline
\{6,10,12,13\}  & 0.78 & \{1,2,3,9,10,11,12,13\} & 0.60 \\ \hline
\end{tabular}
\end{center}
\end{table}

\vspace{-0.4cm}
\section{CONCLUSION}
In this paper, we proposed a new problem -- novelty detection in the MIML setting
and offered a framework based on score functions to solve the problem.
A large number of simulations show that our algorithm not only works well on
synthetic data but also on real-world data. We also demonstrate that
the presence of unlabeled examples in the training set is useful to detect new class examples while testing.
We present the advantage in the MIML setting for novelty detection.
Even though positive examples for the novelty that are not directly labeled,
their presence provides a clear advantage over methods that rely on data that does not include novel class examples.

There are many relative problems call for investigation.
One will be on how to use the information of bag-level labels in detection
if bag-level labels are available, which will possibly improve the performance of our algorithm since
we did not make use of such information in our experiment.

\bibliographystyle{IEEEbib}
\bibliography{mlsp2013}

\end{document}